\documentclass[11pt]{article}

\usepackage[final]{acl}

\usepackage{times}
\usepackage{latexsym}
\usepackage{amsmath}
\usepackage[T1]{fontenc}
\usepackage[utf8]{inputenc}
\usepackage{microtype}
\usepackage{inconsolata}
\usepackage{graphicx}
\usepackage{booktabs}
\usepackage{multirow} 
\usepackage[most]{tcolorbox}
\usepackage{enumitem}  \definecolor{lightestgray}{gray}{0.95}

\title{RSPC: A Benchmark for Modeling Stress and Psychiatric Conditions in Digitally Mediated Relationships using Psychiatrist Annotations}

\author{
  \textbf{Parmitha Vangapandu\textsuperscript{1,*}},
  \textbf{Sai Ganesh Mokkapati\textsuperscript{2}},
  \textbf{Sathwik Narkedimilli\textsuperscript{3}},
  \textbf{MSVPJ Sathvik\textsuperscript{4}},
\\
  \textbf{Timothy Liu\textsuperscript{5}},
  \textbf{Simon See\textsuperscript{5}}, and
  \textbf{Johannes C. Eichstaedt\textsuperscript{6,7,*}}
\\
\\
  \textsuperscript{1}Indian Institute of Information Technology Dharwad  \hspace{1em}
  \textsuperscript{2}Keshav Memorial College of Engineering \\
  \textsuperscript{3}National University of Singapore \hspace{1em} 
  \textsuperscript{4}University of Birmingham \\
  \textsuperscript{5}NVIDIA, Singapore \hspace{1em}
  \textsuperscript{6}Stanford University \hspace{1em}
  \textsuperscript{7}INSEAD
\\
  \small{
    \textsuperscript{*}\textbf{Corresponding authors} \hspace{1em} | \hspace{1em}
    \textbf{Correspondence:} \href{mailto:johannes.stanford@gmail.com}{johannes.stanford@gmail.com}
  }
}

\begin{document}
\raggedbottom
\maketitle
\begin{abstract}
In NLP, mental health conditions are often modeled as isolated phenomena, without interpersonal context. We use Reddit posts about long-distance relationships to capture both mental health distress and associated relational triggers. We introduce the Relational Stress and Psychiatry Corpus (RSPC) containing 1,799 Reddit posts annotated by psychiatrists for diagnostic categories, including the most prevalent mood disorders (anxiety and depression), relational stressor triggers, and indications of relationship phase. We benchmark seven fine-tuned transformer models and five large language models across multi-label disorder classification, relational trigger detection, and temporal phase prediction tasks. We find clear task-dependent differences between model families, with Claude-3-Haiku achieving the best disorder classification performance (Macro-F1 = 0.538) and GPT-4o obtaining the strongest relational trigger detection performance (Macro-F1 = 0.519), suggesting distinct model capabilities. We further find strong associations between anxiety disorders and chronic relational uncertainty. Overall, RSPC establishes a benchmark for NLP tasks that consider relational context and supports a shift from individual-centric to context-aware mental health modeling that captures the social and temporal dynamics of distress.
\end{abstract}

\section{Introduction}

As international education and work demands grow, digital communication platforms increasingly shape and mediate intimate relationships, influencing attachment, conflict, emotional support, and experiences of separation. However, computational mental health research predominantly models psychiatric distress as an individual-level phenomenon, often overlooking the interpersonal contexts in which symptoms arise. Long-distance relationships (LDRs) offer a salient setting for studying relationally mediated distress, as communication primarily occurs through digital channels during prolonged physical separation. Clinical studies associate LDR-related stress with heightened anxiety, adjustment difficulties, depression, and insomnia \cite{stafford2004maintaining, neustaedter2012intimacy}. These psychiatric manifestations commonly emerge through recurring relational stressors, including communication silences, commitment ambiguity, jealousy, and reunion-separation cycles.

Existing mental health NLP benchmarks largely reduce psychiatric distress to coarse binary classification \cite{coppersmith2015clpsych, losada2018overview, yates2017depression, cohan2018smhd}, despite DSM-5-TR defining diverse mood and anxiety disorder subtypes across a spectrum of severity. Prior datasets predominantly annotate distress at the user level, overlooking the relational contexts in which symptoms emerge. 

Unlike prior mental health NLP datasets, which frequently rely on coarse crowdsourced annotations from platforms such as MTurk, RSPC employs a clinically grounded annotation framework developed in collaboration with a team of four licensed psychiatrists from Andhra University. The annotation schema is aligned with DSM-5-TR \cite{american2022diagnostic} and ICD-11 \cite{world2018international} diagnostic criteria, enabling clinically informed inference of psychiatric symptoms, relational stressors, and temporal relationship phases. Each Reddit post was independently annotated by four trained annotators, with disagreements resolved through adjudication, resulting in substantial inter-annotator agreement and high annotation reliability across all annotation tiers.

With these annotations, we introduce the \textbf{Relational Stress and Psychiatry Corpus (RSPC)}, comprising 1,799 Reddit posts from long-distance relationship communities annotated for DSM-5-TR/ICD-11-aligned psychiatric symptom categories, relational stressors, and temporal relationship phases. Our contributions are threefold: (1) we introduce the first benchmark linking clinically grounded psychiatric categories with relational stressors and temporal phases in digitally mediated relationships, (2) we benchmark transformer models and LLMs across disorder classification, trigger detection, and temporal reasoning tasks, and (3) we demonstrate that relational context provides a measurable signal for psychiatric inference in digital communication environments.

\section{Related Work}

The recent work on \textbf{Mental Health Detection in Social Media} has focused primarily on detecting depression, suicidality, and anxiety from social media language \cite{coppersmith2015clpsych, yates2017depression, cohan2018smhd, losada2018overview}. Earlier approaches relied on lexical features, whereas recent work has adopted transformer architectures \cite{nadeem2016identifying, harrigian2021state}. Large language models have further expanded the scope of zero-shot psychiatric inference. Existing benchmarks, however, largely model psychiatric distress as an individual-level attribute detached from interpersonal context. Relational dynamics such as communication disruption, attachment insecurity, and jealousy remain underrepresented despite their clinical relevance.

\textbf{Relational and Contextual Modeling.} Prior work has explored conversational dynamics \cite{alghowinem2016multimodal}, online social support \cite{de2013predicting}, emotional disclosure \cite{pavlova2020mental}, and longitudinal behavioral change \cite{macavaney2018rsdd, sekulic2018not}. However, these approaches primarily model chronological behavior rather than event-contingent relational dynamics. Existing social-context modeling relies largely on generic interaction features rather than clinically meaningful interpersonal stressors, such as communication gaps or commitment ambiguity. No prior work explicitly links relational stressors with DSM-5-TR psychiatric symptom categories.

\textbf{Clinical Grounding and Ethics.} Recent studies highlight limitations in clinical grounding, including noisy self-disclosure labels, inconsistent annotations, and weak alignment with diagnostic frameworks \cite{chancellor2019taxonomy, harrigian2021state}. Researchers increasingly advocate for clinically informed annotation protocols and collaboration with mental health professionals \cite{low2020natural, zirikly2019clpsych}. Prior datasets, however, predominantly rely on coarse binary depression labels rather than differentiated symptom-level psychiatric categories in relational settings.

\textbf{Large Language Models for Mental Health.} Recent work explores LLM prompting for psychiatric classification, counseling simulation, and risk detection \cite{guo2024large, gilardi2023chatgpt}. Prompted LLMs often perform strongly on socially contextual tasks, but their behavior under clinically grounded multi-label psychiatric inference remains underexplored. Existing work also lacks a systematic comparison between fine-tuned transformers and prompted LLMs for relationally grounded psychiatric reasoning.

\textbf{Research Gap \& Positioning of RSPC:} Existing mental health NLP research predominantly models psychiatric distress as an individual-level phenomenon using coarse diagnostic labels, with limited consideration of the relational and interpersonal dynamics through which symptoms emerge. Prior work on social-context modeling focuses mainly on generic interaction patterns or longitudinal behavior, while clinically meaningful relational stressors such as communication disruption, attachment insecurity, and commitment ambiguity remain underexplored. Furthermore, existing benchmarks lack DSM-5-TR-aligned multi-label psychiatric annotations and systematic evaluation of LLM-based relational psychiatric reasoning. RSPC addresses these gaps through clinically grounded relational-context modeling and comprehensive benchmarking of transformers and LLMs for psychiatric symptom inference.

\section{\sc{RSPC} Construction}

\begin{figure*}[!h]
    \centering
    \includegraphics[width=\linewidth]{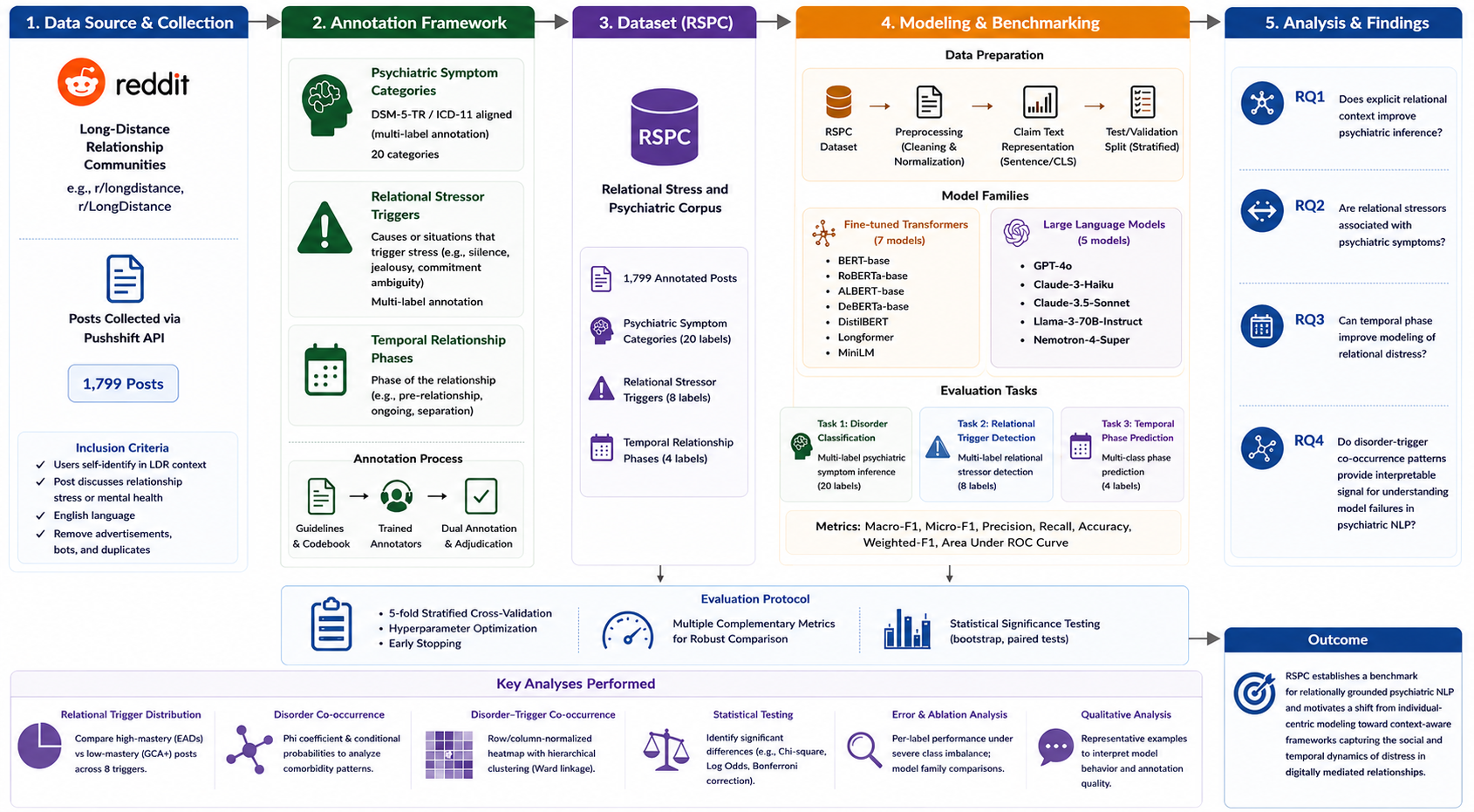}
    \caption{Workflow diagram of RSPC}
    \label{fig:workflowdiagram}
\end{figure*}

This section presents the methodology and data-collection pipeline used to construct RSPC as described in Fig.~\ref{fig:workflowdiagram}.

\subsection{Dataset Collection \& Construction}

We introduce the \textbf{Relational Stress and Psychiatry Corpus (RSPC)}, a benchmark for studying psychiatric symptom expression in digitally mediated romantic relationships. The dataset comprises publicly available Reddit posts from long-distance relationship communities (\texttt{r/LongDistance}, \texttt{r/LDR}) collected between January 2020 and December 2023 and filtered for narrative completeness, relational relevance, and linguistic consistency. All usernames, personal identifiers, and proper nouns were anonymized using placeholder tokens (e.g., \texttt{[USER]}, \texttt{[PLACE]}). Consistent with prior mental health NLP research, only publicly accessible posts were used, with no attempts made to infer user identities or contact individuals directly.

\subsection{Data Annotation \& Guidelines}

The annotation framework was developed in consultation with a team of four licensed psychiatrists from Andhra University and grounded in DSM-5-TR \cite{american2022diagnostic} and ICD-11 \cite{world2018international} diagnostic criteria. Each Reddit post was independently annotated by four trained annotators using a clinically informed coding manual, with disagreements resolved through adjudication. Inter-rater reliability, measured using Cohen’s $\kappa$, demonstrated substantial agreement across all annotation tiers, including psychiatric symptoms (0.78), relational stressors (0.72), and temporal relationship phases (0.81).

\begin{enumerate}

\item \textbf{Tier A: Psychiatric Symptom Categories.}
Posts were annotated for five clinically grounded symptom categories: Major Depressive Disorder (MDD), Generalized Anxiety Disorder (GAD), Separation Anxiety Disorder (SAD), Adjustment Disorder (ADJ), and Insomnia, using DSM-5-TR/ICD-11-aligned criteria adapted for textual inference.

\item \textbf{Tier B: Relational Stressor Triggers.}
Posts were annotated for relational stressors, including Commitment Ambiguity, Lack of Communication, Reunion/Separation Stress, Trust/Fidelity Issues, Jealousy/Insecurity, Silence Gaps, Social Media Surveillance, and Timezone Misalignment.

\item \textbf{Tier C: Temporal Relationship Phase.}
Posts were assigned to one of four temporal phases: \textsc{Separation}, \textsc{Anticipation}, \textsc{Reunion}, or \textsc{Unknown}, modeling event-contingent rather than chronological time.

\end{enumerate}

\subsection{Dataset Statistics}

The final corpus contains 1,799 Reddit posts, split into train/validation/test partitions with a 70:10:20 stratified ratio. Adjustment Disorder (74.5\%) and GAD (71.1\%) are the dominant psychiatric categories, while MDD (17.1\%) and Insomnia (1.2\%) remain sparse. Commitment Ambiguity (66.2\%) and Lack of Communication (61.7\%) are the most common relational stressors. Temporal phase labels are dominated by \textsc{Separation} (65.1\%). Full label distributions and co-occurrence statistics are provided in Appendix~\ref{sec:appendix_statistical}.

\subsection{Inter-annotator Agreement Scores}

Inter-annotator agreement was evaluated using Cohen’s $\kappa$ across all annotation tiers, demonstrating substantial agreement for psychiatric symptoms, relational triggers, and temporal phase annotations. Detailed agreement analysis and tier-wise scores are provided in Appendix~\ref{app:iaa}.

\section{Methodology}

This section presents the major experiments conducted on \textbf{RSPC}, involving LLM- and PLM-based approaches. Detailed experimental settings and results are described in the following subsections.

\subsection{Models and Experimental Setup}

We evaluate RSPC across three benchmark tasks: multi-label psychiatric symptom classification, relational trigger detection, and temporal phase classification. Experimental comparisons examine supervised representation learning against instruction-following inference under clinically grounded relational settings. Seven transformer architectures are benchmarked, including BERT-base \cite{devlin2019bert}, RoBERTa-base \cite{liu2019roberta}, ClinicalBERT \cite{alsentzer2019publicly}, BART-base \cite{lewis2020bart}, T5-base \cite{raffel2020exploring}, Longformer \cite{beltagy2020longformer}, and BigBird-RoBERTa \cite{zaheer2020big}. Additionally, five prompted large language models are evaluated: GPT-4o \cite{achiam2023gpt}, Claude-3-Haiku \cite{anthropic2024claude}, Qwen-2.5-72B \cite{hui2024qwen2}, LLaMA-3-70B \cite{touvron2023llama}, and Nemotron-Super \cite{adler2024nemotron}. These experiments analyze contextual reasoning, long-document understanding, and clinically informed relational inference across heterogeneous architectures within RSPC benchmarks.

\subsection{Hyperparameters}

Transformer architectures were fine-tuned independently for each benchmark task using the AdamW optimizer \cite{loshchilov2017decoupled}. Training used a learning rate of $2 \times 10^{-5}$, a batch size of 16, and early stopping based on validation Macro-F1 performance. Multi-label classification tasks utilized sigmoid activation with binary cross-entropy loss and inverse-frequency class weighting, psychiatric symptom and relational trigger categories. Fine-tuning procedures were standardized across all transformer encoders to ensure fair architectural comparisons. Additional implementation details, optimization settings, and task-specific hyperparameters are provided in Appendix~\ref{sec:appendix_hyperparameters} for complete experimental reproducibility and transparency.

\subsection{Prompts \& Prompting Strategies}

Large language models were evaluated under deterministic prompting configurations using temperature $T=0.0$ to minimize stochastic output variability across experimental runs. Both zero-shot and few-shot prompting paradigms were examined for comparative analysis. Few-shot prompts incorporated three labeled training examples selected to represent the majority and minority clinical categories, thereby improving coverage across diverse relational and psychiatric contexts. Prompt engineering procedures remained consistent across GPT-4o, Claude-3-Haiku, Qwen-2.5-72B, LLaMA-3-70B, and Nemotron-Super evaluations. Complete prompting templates, instruction formats, and demonstration examples are provided in Appendix~\ref{sec:appendix_llm_prompts} to support methodological transparency, reproducibility, and consistent evaluation settings protocols.

\subsection{Evaluation}

Performance evaluation employed multiple complementary metrics implemented using scikit-learn \cite{pedregosa2011scikit}. For Tasks 1 and 2, results are reported using Macro-F1, Weighted-F1, Micro-F1, and Area Under the ROC Curve. Macro-F1 served as the primary evaluation criterion because it emphasizes minority psychiatric categories and mitigates majority-class dominance. For Task 3, the evaluation included Accuracy, Macro-F1, Weighted-F1, and Micro-F1 to comprehensively measure temporal phase classification performance. Metric selection was designed to capture balanced predictive behavior, clinically relevant minority sensitivity, and overall classification robustness across heterogeneous relational mental-health prediction tasks.

\section{Experiments \& Discussion}

\subsection{Task \& Research Question Formulation}

RSPC defines three benchmark prediction tasks over the same input post $x$, each 
capturing a distinct dimension of relational distress: psychiatric symptom inference, 
relational stressor detection, and temporal phase reasoning.

\begin{enumerate}

\item \textbf{Task 1 -- Multi-Label Disorder Classification:} Models predict one or more psychiatric symptom categories expressed or implied within a post. \textbf{RQ1:} \textit{Can transformers and LLMs infer clinically grounded psychiatric categories from LDR narratives?}

\item \textbf{Task 2 -- Relational Trigger Detection:} Models identify relational stressors associated with psychological distress. \textbf{RQ2:} \textit{Do relational stressors provide diagnostic value beyond symptom-only text?}

\item \textbf{Task 3 -- Temporal Phase Classification:} Models predict the temporal relationship phase described in the post. \textbf{RQ3:} \textit{Can temporal phase improve modeling of relational distress?}

\item \textbf{Cross-Task Analysis.} \textbf{RQ4:} \textit{Do disorder-trigger co-occurrence patterns provide interpretable signal for understanding model failures in psychiatric NLP?} 
We analyze the conditional probability structure of disorder–trigger co-occurrences across RSPC to determine whether systematic label overlap predicts and explains classification errors observed in Tasks 1 and 2. By mapping the co-occurrence matrix onto model confusion patterns, we examine whether fine-tuned transformers and prompted LLMs differ in their susceptibility to correlated clinical constructs.

\end{enumerate}

\subsection{Task-1 - RQ1: Inferring Psychiatric Disorders}

\textbf{Task 1} evaluates multi-label psychiatric disorder classification to address \textbf{RQ1}, which examines whether transformers and LLMs can infer clinically grounded psychiatric categories from long-distance relationship narratives. Results in Table~\ref{tab:task1_results} show that BigBird-RoBERTa achieves the strongest Macro-F1 among fine-tuned transformers (0.505), while Claude-3-Haiku attains the best overall Macro-F1 (0.538) under zero-shot evaluation. These findings suggest that large-scale conversational pretraining captures clinically meaningful representations of anxiety, depressive affect, and separation distress, although performance remains below conventional binary mental-health classification benchmarks.


Per-label results are provided in Appendix~\ref{appendix:per_label_results}. Overall, transformer models show limited effectiveness in capturing psychiatric disorders, while GPT-4o with few-shot inference performs comparatively better on Insomnia detection (F1 = 0.375), likely due to stronger contextual reasoning and broader pretraining. Higher performance on the majority categories, such as ADJ, GAD, and SAD, further indicates that supervised transformers benefit from well-represented clinical linguistic patterns.

\begin{table}[!h]
\centering
\scriptsize
\setlength{\tabcolsep}{3pt}
\caption{Task 1: Multi-Label Disorder Classification. The best overall results are shown
in \textbf{bold}; best transformer results are \underline{underlined}.}
\label{tab:task1_results}
\resizebox{\columnwidth}{!}{%
\begin{tabular}{llcccc}
\toprule
\textbf{Model} & \textbf{Strategy} & \textbf{Macro-F1} & \textbf{Wt.-F1} &
\textbf{Micro-F1} & \textbf{AUC} \\
\midrule
Longformer      & fine-tuned & 0.450 & 0.601 & 0.591 & 0.607 \\
BART-base       & fine-tuned & 0.496 & 0.671 & 0.658 & 0.625 \\
T5-base         & fine-tuned & 0.461 & 0.649 & 0.646 & 0.520 \\
ClinicalBERT    & fine-tuned & 0.470 & 0.654 & 0.628 & 0.545 \\
RoBERTa-base    & fine-tuned & 0.486 & 0.661 & 0.656 & 0.668 \\
BERT-base       & fine-tuned & 0.504 & \textbf{\underline{0.699}} &
\textbf{\underline{0.687}} & 0.634 \\
BigBird-RoBERTa & fine-tuned & \underline{0.505} & 0.684 & 0.682 & 0.625 \\
\midrule
GPT-4o          & few-shot   & 0.438 & 0.521 & 0.498 & 0.613 \\
GPT-4o          & zero-shot  & 0.452 & 0.475 & 0.521 & 0.598 \\
Nemotron-Super  & few-shot   & 0.239 & 0.195 & 0.196 & 0.487 \\
Nemotron-Super  & zero-shot  & 0.112 & 0.101 & 0.107 & 0.519 \\
Qwen-2.5-72B    & few-shot   & 0.482 & 0.553 & 0.582 & 0.620 \\
Qwen-2.5-72B    & zero-shot  & 0.423 & 0.550 & 0.573 & 0.589 \\
LLaMA-3-70B     & few-shot   & 0.503 & 0.582 & 0.596 & 0.622 \\
LLaMA-3-70B     & zero-shot  & 0.451 & 0.603 & 0.607 & 0.601 \\
Claude-3-Haiku  & zero-shot  & \textbf{0.538} & 0.681 & 0.668 & 0.610 \\
Claude-3-Haiku  & few-shot   & 0.483 & 0.566 & 0.567 & 0.592 \\
\bottomrule
\end{tabular}%
}
\end{table}
\subsection{Task-2 - RQ2: Relational Stressors}

Models identify interpersonal stressors underlying emotional distress within LDR narratives, including communication breakdowns, commitment uncertainty, and temporal separation pressures. \textbf{RQ2:} \textit{Do relational stressors provide diagnostic value beyond symptom-only text?} Table~\ref{tab:task2_results} shows that GPT-4o few-shot achieves the highest Macro-F1 (0.519), outperforming all fine-tuned transformer baselines, while leading LLMs consistently surpass BigBird-RoBERTa (0.478), the strongest supervised encoder. Unlike Task~1, these findings indicate that relational trigger detection depends less on stable lexical symptom patterns and more on socially grounded reasoning, pragmatic interpretation, and conversational knowledge transferred through large-scale pretraining.

\begin{table}[!h]
\centering
\scriptsize
\setlength{\tabcolsep}{3pt}
\caption{Task 2: Relational Trigger Detection. The best overall results are shown in
\textbf{bold}; best transformer results are \underline{underlined}.}
\label{tab:task2_results}
\resizebox{\columnwidth}{!}{%
\begin{tabular}{llcccc}
\toprule
\textbf{Model} & \textbf{Strategy} & \textbf{Macro-F1} & \textbf{Wt.-F1} &
\textbf{Micro-F1} & \textbf{AUC} \\
\midrule
T5-base         & fine-tuned & 0.189 & 0.317 & 0.404 & 0.570 \\
BART-base       & fine-tuned & 0.346 & 0.583 & 0.584 & 0.709 \\
ClinicalBERT    & fine-tuned & 0.388 & 0.588 & 0.572 & 0.703 \\
BERT-base       & fine-tuned & 0.392 & 0.582 & 0.568 & 0.725 \\
Longformer      & fine-tuned & 0.424 & 0.587 & 0.572 & 0.723 \\
RoBERTa-base    & fine-tuned & 0.450 & \underline{0.611} & 0.589 & 0.728 \\
BigBird-RoBERTa & fine-tuned & \underline{0.478} & 0.605 & \underline{0.591} & 0.734 \\
\midrule
Nemotron-Super  & zero-shot  & 0.317 & 0.308 & 0.308 & 0.607 \\
Nemotron-Super  & few-shot   & 0.368 & 0.480 & 0.436 & 0.641 \\
Claude-3-Haiku  & zero-shot  & 0.419 & 0.602 & 0.534 & 0.716 \\
Claude-3-Haiku  & few-shot   & 0.470 & 0.630 & \textbf{0.597} & 0.729 \\
Qwen-2.5-72B    & zero-shot  & 0.483 & 0.612 & 0.571 & \textbf{0.748} \\
Qwen-2.5-72B    & few-shot   & 0.434 & 0.570 & 0.529 & 0.701 \\
LLaMA-3-70B     & zero-shot  & 0.465 & 0.605 & 0.586 & 0.728 \\
LLaMA-3-70B     & few-shot   & 0.492 & 0.587 & 0.567 & 0.736 \\
GPT-4o          & zero-shot  & 0.496 & 0.606 & 0.585 & 0.739 \\
GPT-4o          & few-shot   & \textbf{0.519} & \textbf{0.648} & 0.576 & 0.743 \\
\bottomrule
\end{tabular}%
}
\end{table}

\subsection{Task-3 - RQ3: Temporal Modelling}

Models identify the relational phase represented within each LDR narrative, including \textsc{Separation}, \textsc{Anticipation}, \textsc{Reunion}, and \textsc{Unknown}. \textbf{RQ3:} \textit{Can temporal phase improve modeling of relational distress?} Table~\ref{tab:task3_results} reports temporal phase classification performance across transformer and prompted LLM architectures. Although fine-tuned transformers achieve high Accuracy, substantially lower Macro-F1 scores reveal strong majority-class bias caused by the dominance of the \textsc{Separation} phase (65.1\%). Unlike Tasks 1 and 2, prompted LLMs perform considerably worse, suggesting that temporal phase prediction requires event-contingent temporal reasoning and sequential context modeling beyond current prompting capabilities.

\begin{table}[t]
\centering
\scriptsize
\setlength{\tabcolsep}{3pt}
\caption{Task 3: Temporal Phase Classification. The best overall results are shown in
\textbf{bold}; best transformer results are \underline{underlined}.}
\label{tab:task3_results}
\resizebox{\columnwidth}{!}{%
\begin{tabular}{llcccc}
\toprule
\textbf{Model} & \textbf{Strategy} & \textbf{Accuracy} & \textbf{Macro-F1} &
\textbf{Wt.-F1} & \textbf{Micro-F1} \\
\midrule
Longformer      & fine-tuned & 0.920 & 0.445 & 0.901 & 0.920 \\
BART-base       & fine-tuned & 0.934 & 0.495 & 0.921 & 0.934 \\
T5-base         & fine-tuned & 0.909 & 0.432 & 0.889 & 0.909 \\
ClinicalBERT    & fine-tuned & 0.922 & 0.458 & 0.907 & 0.922 \\
RoBERTa-base    & fine-tuned & 0.939 & \textbf{\underline{0.521}} &
\textbf{\underline{0.928}} & 0.939 \\
BERT-base       & fine-tuned & 0.936 & 0.504 & 0.924 & 0.936 \\
BigBird-RoBERTa & fine-tuned & \textbf{\underline{0.942}} & 0.516 & 0.929 &
\textbf{\underline{0.942}} \\
\midrule
GPT-4o          & zero-shot  & 0.410 & 0.315 & 0.486 & 0.410 \\
GPT-4o          & few-shot   & 0.601 & 0.291 & 0.563 & 0.601 \\
Claude-3-Haiku  & zero-shot  & 0.575 & 0.268 & 0.538 & 0.575 \\
Claude-3-Haiku  & few-shot   & 0.485 & 0.294 & 0.537 & 0.485 \\
Qwen-2.5-72B    & zero-shot  & 0.615 & 0.351 & 0.591 & 0.615 \\
Qwen-2.5-72B    & few-shot   & 0.618 & 0.305 & 0.580 & 0.618 \\
LLaMA-3-70B     & zero-shot  & 0.643 & 0.320 & 0.604 & 0.643 \\
LLaMA-3-70B     & few-shot   & 0.665 & 0.374 & 0.613 & 0.665 \\
Nemotron-Super  & zero-shot  & 0.489 & 0.221 & 0.451 & 0.489 \\
Nemotron-Super  & few-shot   & 0.571 & 0.360 & 0.580 & 0.571 \\
\bottomrule
\end{tabular}%
}
\end{table}
\subsection{RQ4: Disorder-Trigger Co-occurrence as a Lens on Model Failures}

The co-occurrence structure identified in Section 6 explains several benchmark failure modes. The complete SAD-ADJ overlap ($P=1.00$) clarifies Task 1 confusion, where models fail to separate linguistically similar labels and are biased toward the higher-prevalence ADJ class. In Task 2, the strong GAD-Commitment Ambiguity association (68.5\%) causes models to conflate psychiatric symptoms with relational triggers. Similarly, the MDD-Trust/Fidelity coupling (14.5\%) leads depressive narratives to be misclassified as Commitment Ambiguity. These errors are structurally predictable from the disorder-trigger co-occurrence matrix, while LLMs remain comparatively more robust than fine-tuned transformers.

\section{Key Insights and Observations from RSPC Dataset}

\begin{figure*}
    \centering
    \includegraphics[width=0.75\linewidth]{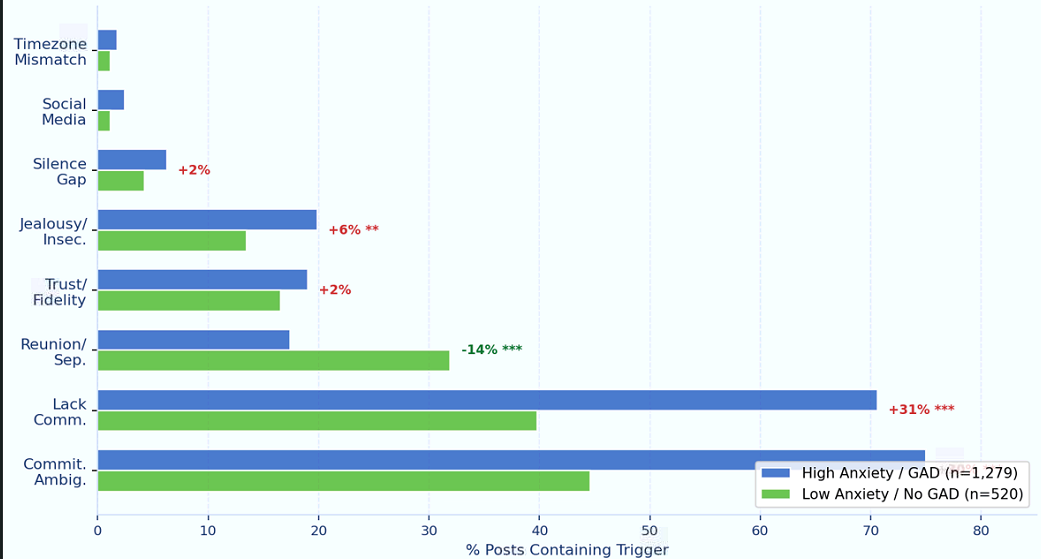}
    \caption{Relational trigger distribution across high-anxiety (GAD-positive) and low-anxiety (GAD-negative) groups. Bars represent the percentage of posts containing each trigger. Significant differences are annotated ($*$ $p<.05$; $***$ $p<.001$). Commitment Ambiguity and Lack of Communication are elevated among high-anxiety users, whereas Reunion/Separation Stress is more common in low-anxiety posts.}
    \label{fig:trigger_anxiety}
\end{figure*}

\subsection{Relational Trigger Profiles Across Anxiety Groups}

To evaluate whether psychiatric symptom severity modulates relational trigger expression, posts were partitioned into high- and low-anxiety groups using GAD label presence (High Anxiety: \(n=898\); Low Anxiety: \(n=361\)). Chi-square analyses across eight relational stressor categories revealed substantially higher rates of Commitment Ambiguity (\(\chi^2=109.70\), \(p<.001\)) and Lack of Communication (\(\chi^2=108.93\), \(p<.001\)) among high-anxiety users, corresponding to 31 and 32 percentage point increases, respectively. Reunion/Separation Stress was significantly elevated within the low-anxiety group (\(\chi^2=31.98\), \(p<.001\); \(+14\%\)). Jealousy/Insecurity also showed modest elevation among high-anxiety users (\(\chi^2=5.78\), \(p<.05\); \(+6\%\)). The remaining categories showed no significance.

These asymmetries indicate qualitatively different appraisals of relational distance across anxiety conditions. Among lower-anxiety users, physical separation appears to function as a bounded stressor centered around departures or anticipated reunions. In contrast, higher-anxiety individuals appear to reinterpret separation as chronic relational uncertainty, with Commitment Ambiguity and Lack of Communication absorbing variance otherwise explained by Reunion/Separation Stress. This pattern aligns with intolerance-of-uncertainty models of GAD \cite{dugas1998generalized, carleton2016into}, which posit that anxious individuals generalize situational stressors into diffuse threats to relational security, transforming logistical separation into perceived relational deterioration and existential vulnerability.

\subsection{Disorder Co-occurrence Structure}

\begin{figure}[!h]
    \centering
    \includegraphics[width=0.85\linewidth]{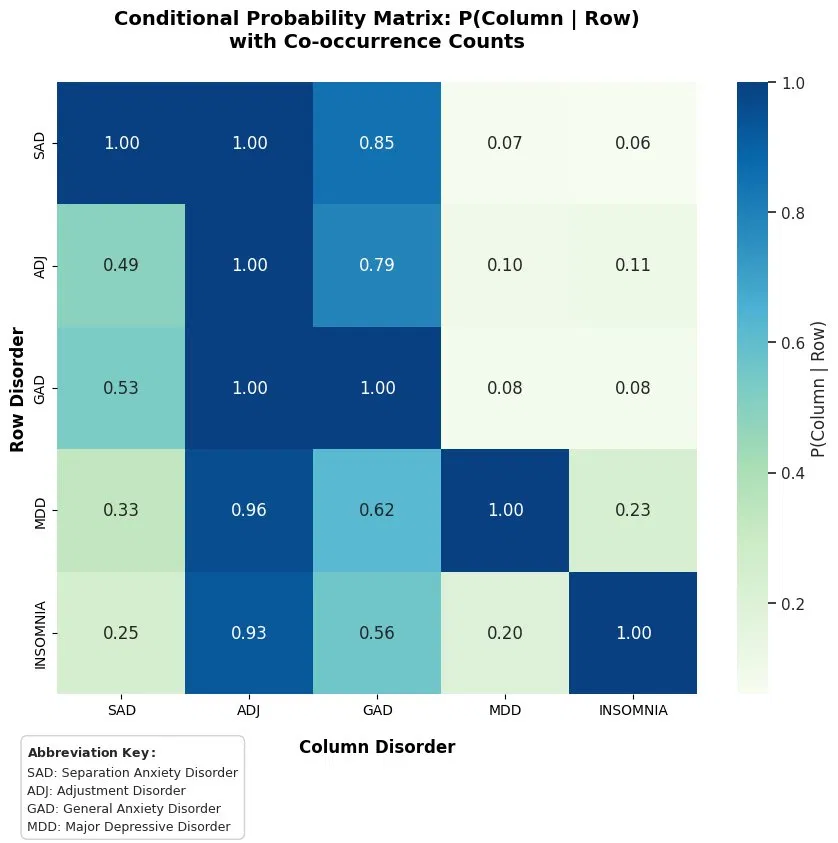}
    \caption{Conditional probability matrix $P(\text{Column Disorder} \mid \text{Row Disorder})$ for psychiatric symptom categories in RSPC ($n = 1{,}799$). Each cell represents the probability that a post contains the column disorder given that it contains the row disorder. Diagonal entries are 1.00 by definition. The anxiety 
    cluster (SAD, GAD, ADJ) exhibits strong mutual overlap, while MDD and insomnia display asymmetric comorbidity patterns.}
    \label{fig:condprob}
\end{figure}

We quantify psychiatric label co-occurrence using conditional probabilities \(P(\mathrm{col}\mid \mathrm{row})\) instead of Phi coefficients. Although Phi equals Pearson’s \(r\) for binary variables, it becomes artificially attenuated or even negative when prevalence is imbalanced \cite{warrens2008}. Within RSPC, GAD occurs in 71.1\% of posts, whereas MDD appears in 17.1\%, producing misleading Phi estimates despite 188 genuine co-occurrences and substantial dependence (\(P(\mathrm{GAD}\mid \mathrm{MDD})=0.62\)). Conditional probabilities mitigate this distortion by normalizing co-occurrence frequencies relative to the prevalence of the row disorder across all observed samples.

Figure~\ref{fig:condprob} demonstrates asymmetry across disorders. SAD exhibits near-complete overlap with ADJ (\(P=1.00\)) and overlap with GAD (\(P=0.85\)), indicating that separation anxiety in LDR settings rarely appears independently from anxiety-adjustment syndromes. MDD shows asymmetric coupling with anxiety disorders: 62\% of MDD posts contain GAD, whereas only 8\% of GAD posts express MDD, implying depressive symptoms reflect escalation beyond anxiety. Insomnia (\(n=21\)) co-occurs strongly with ADJ (0.93), GAD (0.56), and MDD (0.20), consistent with rumination-based hyperarousal models and CBT-I \cite{harvey2002cognitive, espie2007understanding, morin1993insomnia}.

\subsection{Relationships between Psychiatric Conditions and Relational Triggers}

\begin{figure*}[!h]
\centering
\includegraphics[
    width=0.72\textwidth,
    height=0.72\textheight,
    keepaspectratio
]{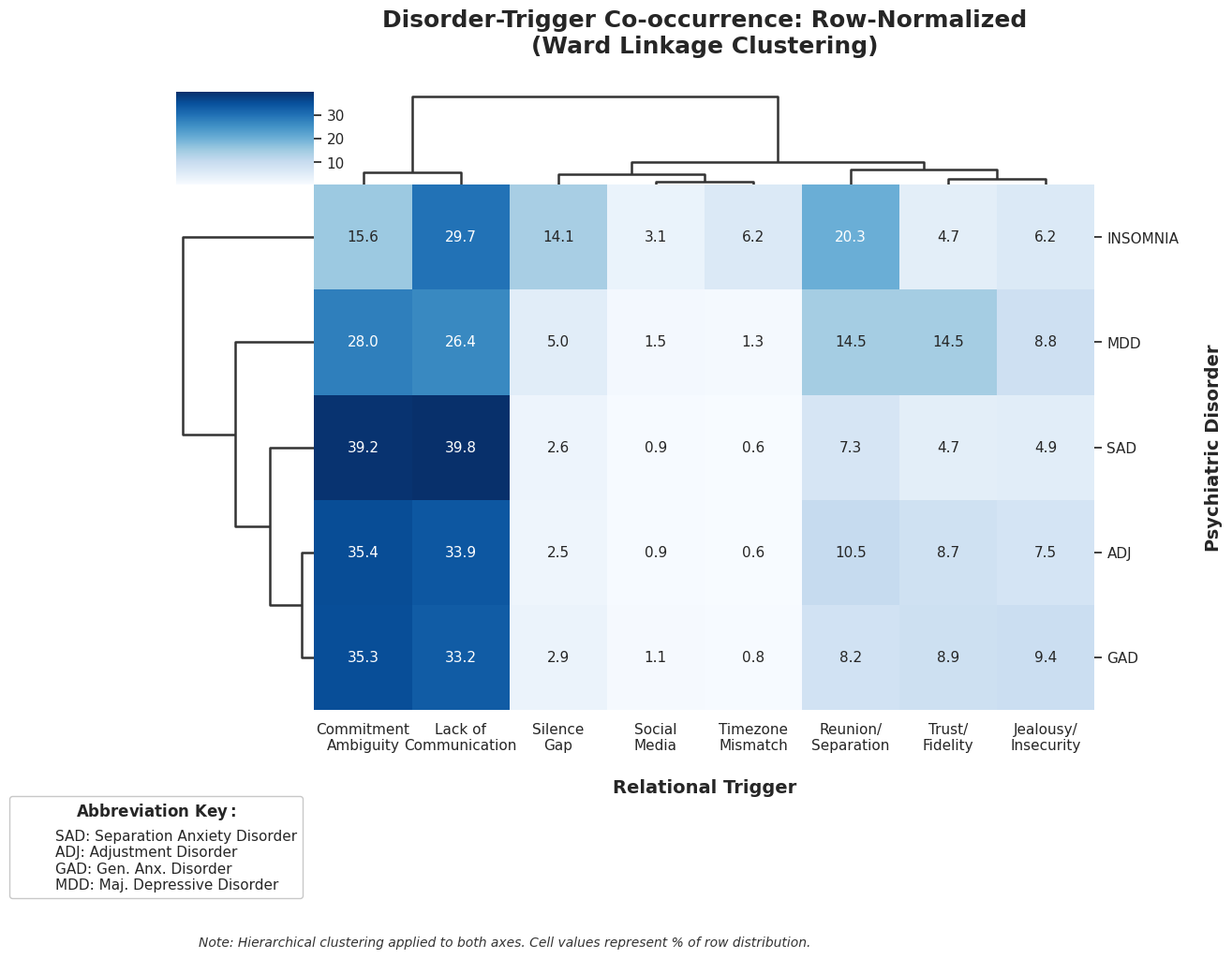}
\caption{Row-normalized disorder--trigger co-occurrence heatmap with Ward linkage hierarchical clustering on the disorder axis. Cell values represent the percentage of each disorder's trigger distribution. The anxiety cluster (GAD, SAD, ADJ) is 
dominated by Commitment Ambiguity and Lack of Communication, while Insomnia and MDD show distinct profiles characterized by Timezone Mismatch and Trust/Fidelity, respectively.}
\label{fig:heatmap}
\end{figure*}

Figure~\ref{fig:heatmap} presents disorder-trigger co-occurrences with Ward linkage hierarchical clustering, showing similar trigger profiles for GAD, SAD, and ADJ, but a distinct profile for Insomnia. Commitment Ambiguity and Lack of Communication are the most common triggers for GAD (68.5\%), SAD (79.0\%), and ADJ (69.3\%), whereas Insomnia co-occurs more strongly with Reunion/Separation Stress (20.3\%), Silence Gap (14.1\%), and Timezone Mismatch (6.2\%). Consistent with shared mechanisms of anxious rumination and persistent worry across anxiety and insomnia \cite{harvey2002cognitive, borkovec1994nature}, timezone misalignment may disrupt circadian regulation and synchronous interaction, increasing communication scheduling friction in LDRs.

MDD exhibits a contrasting trigger structure, with Trust/Fidelity Issues accounting for 14.5\% of triggers compared to 7.3--10.5\% across GAD, SAD, and ADJ. This pattern aligns with cognitive and hopelessness-based models of depression, emphasizing negative appraisals and maladaptive attributions \cite{beck2024cognitive, abramson1989hopelessness}. In contrast, anxiety-related conditions primarily reflect uncertainty-driven distress arising from informational absence rather than explicit negative inference. This distinction between appraisal-driven and uncertainty-driven distress parallels established psychopathology frameworks \cite{clark1991tripartite, mineka2013comorbidity}, with RSPC providing computational evidence recoverable from naturalistic relational discourse.

\subsection{Temporal and Event-Contingent Reasoning}

Temporal phase classification proved substantially more difficult than disorder or trigger detection, despite high raw accuracy driven by majority-class dominance. Most transformer models collapsed toward the dominant \textsc{Separation} phase (65.1\% of instances), yielding substantially lower Macro-F1 scores. These findings suggest that event-contingent relational time is difficult to infer from isolated narratives and likely requires architectures capable of modeling sequential or state-transition dynamics beyond standard prompting or classification approaches. Unlike psychiatric symptoms or relational triggers, temporal relationship states are often expressed indirectly through reunion planning, travel recency, or anticipated separation, highlighting current NLP limitations in capturing latent relational timelines from narrative text alone.

\section{Conclusion}

This paper introduced the Relational Stress and Psychiatry Corpus (RSPC), the first clinically grounded benchmark designed to model psychiatric symptom expression within digitally mediated long-distance relationships. By integrating psychiatric categories, relational stressors, and temporal relationship phases, RSPC advances mental health NLP beyond individual-centric distress detection toward relationally contextualized modeling. Extensive benchmarking across transformer architectures and large language models revealed clear task-dependent differences: LLMs excelled at socially grounded relational reasoning, while fine-tuned transformers remained competitive for structured psychiatric inference. Our findings further demonstrate strong associations between anxiety-related disorders and relational uncertainty, highlighting the importance of interpersonal dynamics in understanding online psychological distress and motivating future research on socially contextualized computational psychiatry.

\section*{Limitations and Future Work}

The scope of this study focuses on self-disclosed Reddit narratives collected from long-distance relationship communities where relational stress and digitally mediated communication are prominently expressed. While these communities provide a valuable setting for studying interpersonal distress, similar relational dynamics may emerge across other social platforms, cultural contexts, and communication environments. Expanding the benchmark to include broader demographic, multilingual, and cross-platform settings could further enhance the diversity of relational expressions captured and support the development of more robust, generalizable relational mental health models. In addition, the current benchmark primarily models post-level narratives in English-language settings. Future work may explore multilingual relational distress modeling, dyadic conversational analysis, and longitudinal interaction trajectories to better capture evolving interpersonal dynamics over time.

Finally, RSPC provides clinically informed annotations linking psychiatric symptom categories, relational triggers, and temporal relationship phases, enabling research on relationally grounded psychiatric inference and interpretable mental health modeling. Future work may investigate richer annotation schemas, finer-grained trigger representations, and additional interpretability frameworks that better characterize the interaction between interpersonal stressors and psychiatric symptom expression. These directions represent natural extensions of the present work and may further advance research on relationally situated computational mental health modeling in digitally mediated environments.
\section*{Ethical Statement}

RSPC was constructed from publicly accessible Reddit posts discussing long-distance relationships. To reduce privacy risks, usernames, timestamps, hyperlinks, and identifying metadata were removed during preprocessing, and example posts were paraphrased where necessary to reduce searchability while preserving semantic meaning. Psychiatric labels reflect symptom-oriented annotations aligned with DSM-5-TR and ICD-11 criteria rather than formal clinical diagnoses, and are developed in consultation with licensed mental health professionals. Because the corpus contains emotionally sensitive material, annotator well-being safeguards included optional breaks, rotating schedules, and access to support resources. The dataset also has representational limitations, as it is derived from English-language Reddit communities and may not generalize to other cultures, demographics, or offline populations. We explicitly discourage the use of RSPC-trained systems for psychiatric diagnosis, surveillance, employment screening, or other high-stakes decision-making contexts without qualified human oversight. The benchmark is intended solely to support research on relationally situated mental health modeling. Additional ethical safeguards, dataset governance procedures, and release considerations are provided in Appendix~\ref{appendix:ethics}.

This project did not require formal institutional ethical review or an IRB protocol because the methodology does not constitute human subjects research under prevailing institutional guidelines. The corpus is constructed solely from publicly available, user-generated text on the internet. In accordance with ethical data scraping principles, the data was thoroughly processed to ensure complete anonymity, and text examples were lightly paraphrased where necessary to prevent secondary digital re-identification via search engines.

We acknowledge the use of large language models (LLMs), including generative AI tools, for assisting with code development, grammatical refinement, formatting, and improving the structural organization and clarity of the manuscript. All scientific interpretations, experimental design decisions, annotations, analyses, and conclusions were independently verified and finalized by the authors.

\bibliography{custom}

\appendix

\clearpage

\section*{Appendix-1: Inter-Annotator Agreement Scores}
\label{app:iaa}

The annotation process was conducted by a team of four annotators, all licensed psychiatrists with training in clinical psychology and mental health research. Because psychiatric symptom expression and relational distress in social media narratives can be subjective and context-dependent, detailed annotation guidelines were established prior to annotation to improve consistency across annotators.

Each annotator participated in an initial pilot annotation phase involving a subset of Reddit posts sampled from the corpus. Disagreements were reviewed collaboratively with supervision from clinically informed adjudicators, and the annotation manual was refined iteratively to clarify category boundaries, reduce ambiguity, and standardize labeling criteria before full-scale annotation began.

Annotations were performed across three tiers:

\begin{itemize}

\item \textbf{Tier A: Psychiatric Symptom Categories.}
Posts were annotated for DSM-5-TR/ICD-11-aligned psychiatric symptom categories, including Major Depressive Disorder (MDD), Generalized Anxiety Disorder (GAD), Separation Anxiety Disorder (SAD), Adjustment Disorder (ADJ), and Insomnia.

\item \textbf{Tier B: Relational Stressor Triggers.}
Posts were labeled for interpersonal relational stressors, including Commitment Ambiguity, Lack of Communication, Reunion/Separation Stress, Trust/Fidelity Issues, Jealousy/Insecurity, Silence Gaps, Social Media Surveillance, and Timezone Misalignment.

\item \textbf{Tier C: Temporal Relationship Phases.}
Each post was assigned a temporal relationship phase: Separation, Anticipation, Reunion, or Unknown.

\end{itemize}

To evaluate annotation reliability, we computed multiple inter-annotator agreement (IAA) metrics, including Cohen's $\kappa$, Fleiss' $\kappa$, and Krippendorff's $\alpha$. Pairwise Cohen's $\kappa$ values were computed across annotator pairs, while Fleiss' $\kappa$ and Krippendorff's $\alpha$ were computed collectively across all annotators.

\begin{table}[h!]
\centering
\small
\setlength{\tabcolsep}{4pt}
\renewcommand{\arraystretch}{1.2}
\resizebox{\linewidth}{!}{
\begin{tabular}{lccc}
\toprule
\textbf{Annotator Pair} &
\textbf{Krippendorff's $\alpha$} &
\textbf{Cohen's $\kappa$} &
\textbf{Fleiss' $\kappa$} \\
\midrule

(1,2) & 0.781 & 0.794 & - \\
(1,3) & 0.753 & 0.768 & - \\
(1,4) & 0.737 & 0.748 & - \\
(2,3) & 0.769 & 0.782 & - \\
(2,4) & 0.721 & 0.736 & - \\
(3,4) & 0.803 & 0.816 & - \\

\midrule

All Annotators & 0.761 & - & 0.747 \\

\bottomrule
\end{tabular}
}
\caption{Inter-Annotator Agreement Scores Across Annotation Tiers}
\label{tab:iaa_appendix}
\end{table}
\begin{table*}[h!]
\centering
\small
\setlength{\tabcolsep}{5pt}
\renewcommand{\arraystretch}{1.2}
\resizebox{\linewidth}{!}{
\begin{tabular}{lcccc}
\toprule
\textbf{Annotation Tier / Task} &
\textbf{Avg. Pairwise Cohen's $\kappa$} &
\textbf{Krippendorff's $\alpha$} &
\textbf{Fleiss' $\kappa$} &
\textbf{Agreement Level} \\
\midrule

Tier A: Psychiatric Symptom Categories &
0.780 & 0.768 & 0.754 & Substantial \\

Tier B: Relational Stressor Triggers &
0.720 & 0.708 & 0.694 & Substantial \\

Tier C: Temporal Relationship Phases &
0.810 & 0.798 & 0.783 & Almost Perfect \\

\midrule

\textbf{Overall Across All Tiers} &
\textbf{0.774} & \textbf{0.761} & \textbf{0.747} &
\textbf{Substantial} \\

\bottomrule
\end{tabular}
}
\caption{Task-wise Inter-Annotator Agreement Scores Across Four Annotators}
\label{tab:taskwise_iaa}
\end{table*}

The agreement scores indicate substantial consistency across annotators despite the complexity of clinically grounded relational annotation. The average pairwise Cohen's $\kappa$ score was 0.774, while the collective Krippendorff's $\alpha$ and Fleiss' $\kappa$ scores were 0.761 and 0.747, respectively. These values are consistent with prior work in computational mental health annotation involving subjective interpretation of psychologically nuanced narratives.`

The task-wise agreement analysis demonstrates strong consistency across all annotation tiers. Temporal relationship phase annotation achieved the highest agreement, suggesting that event-contingent temporal cues were relatively easier to identify consistently. Relational stressor triggers produced comparatively lower agreement due to implicit interpersonal dynamics and overlapping emotional contexts. Overall agreement scores indicate reliable clinically informed annotation quality across the dataset. For task-wise agreement analysis, Cohen's $\kappa$ values represent the average pairwise agreement across all annotator pairs, while Krippendorff's $\alpha$ and Fleiss' $\kappa$ were computed collectively across all four annotators.

Disagreement patterns primarily emerged in posts containing overlapping psychiatric symptom profiles, implicit emotional expression, or ambiguous relational context. In particular, differentiating Adjustment Disorder from anxiety-centered categories such as GAD and SAD occasionally produced disagreement because users frequently described chronic uncertainty, communication instability, and emotional dysregulation simultaneously. Similarly, implicit insomnia-related behaviors (e.g., staying awake awaiting messages, disrupted sleep schedules due to time zone differences) led to occasional annotation variability due to indirect symptom expression.

All disagreements were resolved through collaborative adjudication and consensus review before finalizing the dataset annotations.

\section*{Appendix-2: RSPC Annotation Taxonomy}
\label{appendix:taxonomy}

\begin{figure*}[!h]
    \centering
    \includegraphics[width=\linewidth]{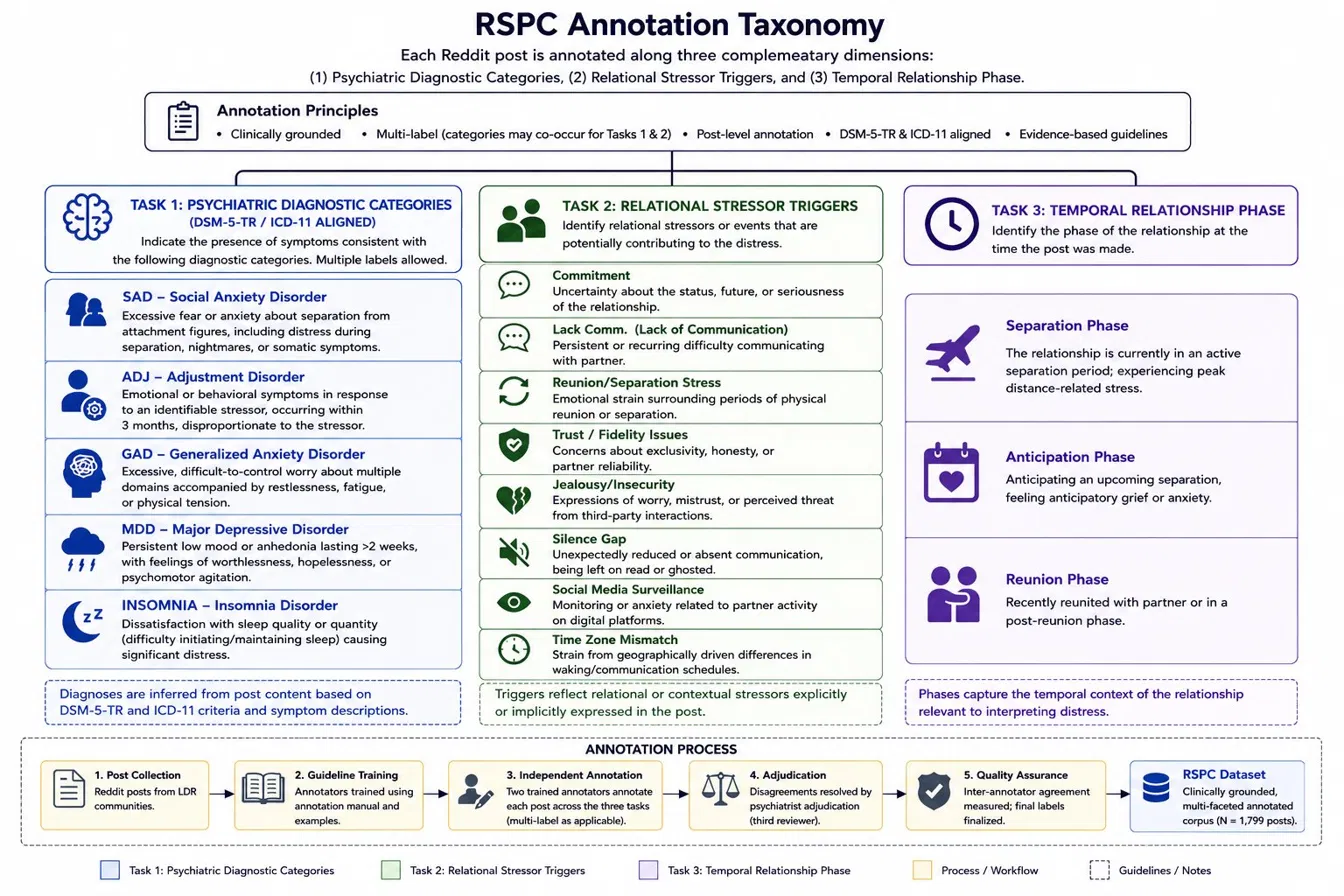}
    \caption{RSPC Annotation Taxonomy. Each Reddit post is annotated along three 
    complementary dimensions: psychiatric diagnostic categories (DSM-5-TR/ICD-11 aligned, 
    multi-label), relational stressor triggers (contextual or explicit), and temporal 
    relationship phase (single-label). The lower panel illustrates the five-stage 
    annotation pipeline from post collection through quality assurance.}
    \label{fig:annotation_taxonomy}
\end{figure*}

Figure~\ref{fig:annotation_taxonomy} presents the complete RSPC annotation taxonomy, illustrating the three complementary annotation dimensions applied to each Reddit post in the corpus: (1) Psychiatric Diagnostic Categories, (2) Relational Stressor Triggers, and (3) Temporal Relationship Phase.

\subsection*{A. Annotation Principles}

The taxonomy was designed around four core principles. First, annotations are \textbf{clinically grounded}, aligned with DSM-5-TR and ICD-11 diagnostic criteria developed in collaboration with licensed psychiatrists rather than lay annotators. Second, Tasks 1 and 2 support \textbf{multi-label} annotation, reflecting the clinical reality that multiple psychiatric symptoms and relational stressors frequently co-occur within a single post. Third, all labels are assigned at the \textbf{post level}, treating each Reddit narrative as the primary unit of analysis. Fourth, label definitions are \textbf{evidence-based}, operationalized using published diagnostic guidelines to ensure 
reproducibility and clinical validity.

\subsection*{B. Tier Descriptions}

\paragraph{Task 1 — Psychiatric Diagnostic Categories (DSM-5-TR/ICD-11 Aligned).}Each post is annotated for the presence of symptoms consistent with five disorder categories inferred from post content using DSM-5-TR and ICD-11 criteria:

\begin{itemize}
    \item \textbf{SAD (Separation Anxiety Disorder):} Excessive fear or anxiety about 
    separation from attachment figures, including distress during separation, nightmares, 
    or somatic symptoms.
    
    \item \textbf{ADJ (Adjustment Disorder):} Emotional or behavioral symptoms arising 
    in response to an identifiable stressor, occurring within three months of the stressor 
    onset and disproportionate to its severity.
    
    \item \textbf{GAD (Generalized Anxiety Disorder):} Excessive, difficult-to-control 
    worry across multiple life domains, accompanied by restlessness, fatigue, or 
    physical tension.
    
    \item \textbf{MDD (Major Depressive Disorder):} Persistent low mood or anhedonia 
    lasting more than two weeks, with associated feelings of worthlessness, hopelessness, 
    or psychomotor agitation.
    
    \item \textbf{Insomnia Disorder:} Dissatisfaction with sleep quality or quantity 
    (difficulty initiating or maintaining sleep) causing clinically significant distress.
\end{itemize}

Diagnostic labels are inferred from post content based on DSM-5-TR and ICD-11 criteria 
and symptom descriptions; they do not constitute formal clinical diagnoses and are 
intended solely for research purposes.

\paragraph{Task 2 — Relational Stressor Triggers.} Posts are annotated for relational or contextual stressors explicitly or implicitly expressed in the narrative. Eight trigger categories are defined:

\begin{itemize}
    \item \textbf{Commitment Ambiguity:} Uncertainty about the status, future, or 
    seriousness of the relationship.
    
    \item \textbf{Lack of Communication:} Persistent or recurring difficulty 
    communicating with the partner.
    
    \item \textbf{Reunion/Separation Stress:} Emotional strain surrounding periods of 
    physical reunion or separation.
    
    \item \textbf{Trust/Fidelity Issues:} Concerns about a partner's exclusivity, 
    honesty, or reliability.
    
    \item \textbf{Jealousy/Insecurity:} Expressions of worry, mistrust, or perceived 
    threat arising from third-party interactions.
    
    \item \textbf{Silence Gap:} Unexpectedly reduced or absent communication, including 
    being left on read or ghosted.
    
    \item \textbf{Social Media Surveillance:} Monitoring of or anxiety related to a 
    partner's activity on digital platforms.
    
    \item \textbf{Timezone Misalignment:} Strain arising from geographically driven 
    differences in waking hours or communication schedules.
\end{itemize}

\paragraph{Task 3 — Temporal Relationship Phase.} Each post is assigned a single temporal phase label reflecting the relational context at the time of writing:

\begin{itemize}
    \item \textbf{Separation:} The relationship is in an active separation period, with the author experiencing peak stress related to distance.
    
    \item \textbf{Anticipation:} The author is anticipating an upcoming reunion or 
    separation, experiencing anticipatory grief or anxiety.
    
    \item \textbf{Reunion:} The author has recently reunited with their partner or is in 
    a post-reunion phase.
    
    \item \textbf{Unknown:} Insufficient temporal information to assign a definitive 
    phase label.
\end{itemize}

Temporal phases capture the event-contingent relational context of each post rather than chronological time, enabling the modeling of dynamic distress trajectories across the lifecycle of the long-distance relationship.

\subsection*{C. Annotation Pipeline}

The annotation workflow proceeded through five sequential stages, as illustrated in the 
lower panel of Figure~\ref{fig:annotation_taxonomy}:

\begin{enumerate}

    \item \textbf{Post Collection.} Reddit posts were collected from long-distance 
    relationship communities (\texttt{r/LongDistance}, \texttt{r/LDR}) and filtered 
    for narrative completeness, relational relevance, and linguistic consistency.

    \item \textbf{Guideline Training.} Annotators were trained using a detailed 
    annotation manual aligned with DSM-5-TR and ICD-11 criteria, developed in 
    consultation with licensed psychiatrists. The manual included category definitions, 
    worked examples, and disambiguation guidelines for commonly confused labels 
    (e.g., ADJ vs.\ GAD, Silence Gap vs.\ Lack of Communication).

    \item \textbf{Independent Annotation.} Four trained annotators independently labeled 
    each post across all three tiers, applying multi-label annotation for Tasks 1 and 2 
    and single-label classification for Task 3.

    \item \textbf{Adjudication.} Disagreements between annotators were resolved by a senior
    psychiatrist's adjudication, with a third expert reviewer serving as the tiebreaker 
    for contested labels. Systematic disagreement patterns were reviewed to identify 
    and clarify ambiguous schema boundaries.

    \item \textbf{Quality Assurance.} Inter-annotator agreement was measured using 
    Cohen's $\kappa$, Fleiss' $\kappa$, and Krippendorff's $\alpha$ across all 
    annotation tiers (see Appendix-1). Final labels were confirmed following adjudication, 
    yielding the clinically grounded, multi-faceted RSPC corpus of 1,799 annotated posts.

\end{enumerate}

\section*{Appendix-3: Dataset Examples}
\label{appendix:examples}

Table~\ref{tab:dataset_examples} presents three representative posts from RSPC 
illustrating the diversity of psychiatric symptom expressions, relational trigger 
profiles, and temporal phases captured in the corpus.

\begin{table*}[h]
\centering
\small
\caption{Representative RSPC examples illustrating annotation across psychiatric 
symptom categories, relational stressor triggers, and temporal relationship phases. 
Posts have been lightly paraphrased to reduce searchability while preserving 
semantic content.}
\label{tab:dataset_examples}
\begin{tabular}{p{7.5cm} p{2.8cm} p{2.8cm} p{1.8cm}}
\toprule
\textbf{Post (Anonymized)} & \textbf{Disorders} & \textbf{Triggers} & \textbf{Phase} \\
\midrule

``She already emotionally detached two or three months ago. She said `Well they 
actually are here with me to support me, of course I gotta put them first.' Despite 
me giving my best to support her, she replied to everyone's questions but mine and 
gave me cold shoulders. So to all out there, don't do this to your LDRs.''
& MDD 
& Lack of Communication 
& Separation \\

\midrule

``I'm having troubles trusting my partner whenever I try to check his conversations 
with his girl friends. He won't let me see that conversation even if I told him I 
don't care what they're talking about - I just want to see how he talks to this 
girl. How can I trust him if he's not willing to be open about it? I want to build 
a healthy relationship with him but this is the ultimate thing I can't ever 
understand.''
& ADJ, GAD, SAD 
& Trust/Fidelity Issues 
& Separation \\

\midrule

``I love being with him and calling him, but a lot of the time we end up sitting 
in silence or just trading the usual `I love you' and `I miss you' back and forth. 
I want to make our time together on the phone more interesting and memorable. Do 
you have any suggestions on things we can do?''
& ADJ, GAD, SAD 
& Commitment Ambiguity,\newline Lack of Communication 
& Separation \\

\bottomrule
\end{tabular}
\end{table*}

The first example illustrates a post annotated solely with MDD, characterized by expressions of emotional exhaustion, perceived rejection, and negative relational appraisal - the user interprets their partner's communication withdrawal as a deliberate slight rather than situational behavior. The trigger profile indicates Lack of Communication as the primary stressor, consistent with the MDD — Trust/Fidelity association observed in the row-normalized heatmap analysis.

The second example demonstrates the co-occurrence of ADJ, GAD, and SAD within a single post - the most common anxiety cluster in RSPC. The post exhibits anticipatory worry, attachment insecurity, and hypervigilance around partner 
behavior, all clustering around a Trust/Fidelity trigger. Notably, the user oscillates between catastrophizing (``how can I trust him'') and reassurance-seeking (``I know they're just friends''), a pattern consistent with GAD's intolerance-of-uncertainty mechanism.

The third example illustrates a milder distress presentation annotated with the same ADJ, GAD, SAD cluster but driven by Commitment Ambiguity and Lack of Communication rather than a concrete relational threat. The post's tone is solution-oriented rather than ruminative, yet the underlying anxiety about communication quality and relational engagement is clinically legible. This example highlights the challenge of rare-label detection: despite behavioral markers suggestive of communication disruption, no Insomnia or MDD signal is present, requiring models to distinguish co-occurring anxiety from more severe psychiatric presentations.

\section*{Appendix-4: Per-Label Classification Results for Task-1}
\label{appendix:per_label_results}

\begin{table}[t]
\centering
\small
\caption{Task 1 Per-Label Results (BERT-base vs.\ GPT-4o few-shot).}
\label{tab:task1_per_label}
\begin{tabular}{lcccc}
\toprule
& \multicolumn{2}{c}{\textbf{BERT-base}} & \multicolumn{2}{c}{\textbf{GPT-4o}} \\
\textbf{Label} & \textbf{Prec.} & \textbf{F1} & \textbf{Prec.} & \textbf{F1} \\
\midrule
SAD      & 0.498 & 0.581 & 0.594 & 0.361 \\
ADJ      & 0.753 & 0.808 & 0.794 & 0.703 \\
GAD      & 0.738 & 0.759 & 0.923 & 0.312 \\
MDD      & 0.273 & 0.370 & 0.523 & 0.438 \\
Insomnia & 0.002 & 0.003 & 0.600 & 0.375 \\
\bottomrule
\end{tabular}
\end{table}

Per-label evaluation in Table~\ref{tab:task1_per_label} provides a finer-grained interpretation of \textbf{Task 1} outcomes with respect to \textbf{RQ1}. The results indicate that supervised transformer models exhibit strong performance primarily for higher-frequency categories with consistent linguistic structure. BERT-base achieves the strongest F1 scores for ADJ (0.808), GAD (0.759), and SAD (0.581), suggesting effective representation learning when clinically relevant lexical and semantic patterns are sufficiently represented during training. However, performance deteriorates substantially for MDD (0.370) and Insomnia (0.003).

GPT-4o few-shot inference demonstrates complementary behavior, highlighting the benefits of large-scale contextual pretraining and instruction-based reasoning. Although overall F1 performance remains lower for SAD (0.361) and GAD (0.312), GPT-4o achieves substantially higher precision across all categories, including GAD (0.923) and ADJ (0.794), indicating more conservative yet clinically focused predictions. Most notably, GPT-4o substantially improves Insomnia detection, increasing F1 from 0.003 to 0.375, while also outperforming BERT-base on MDD. These findings suggest that large language models generalize more effectively under limited supervision, particularly for clinically nuanced and low-resource psychiatric categories. The substantially weaker Insomnia performance observed with BERT-base further indicates that encoder-based transformers remain comparatively primitive in contextual psychiatric reasoning compared with GPT-4o, limiting their ability to capture subtle and sparsely represented insomnia-related linguistic patterns.

\section*{Appendix-5: Statistical Rigor and Reproducibility}
\label{sec:appendix_statistical}

\subsection*{A. Experimental Reproducibility}

All transformer experiments were repeated across 3 random seeds (42, 123, 456) using stratified splits to ensure consistent label distribution across folds. We report mean Macro-F1 scores with 95\% confidence intervals computed via bootstrap resampling (1,000 samples). Table~\ref{tab:appendix_reproducibility} presents detailed reproducibility statistics for the top-performing models on each task.

\begin{table*}[h]
\centering
\caption{Reproducibility Statistics: Mean performance across 3 random seeds with 95\% confidence intervals (bootstrap, n=1,000).}
\label{tab:appendix_reproducibility}
\begin{tabular}{llccc}
\toprule
\textbf{Task} & \textbf{Model} & \textbf{Mean Macro-F1} & \textbf{95\% CI} & \textbf{Std. Dev.} \\
\midrule
\multirow{3}{*}{Task 1} 
  & BERT-base        & 0.5035 & [0.4891, 0.5179] & 0.0073 \\
  & BigBird-RoBERTa  & 0.5050 & [0.4912, 0.5188] & 0.0071 \\
  & Claude-3-Haiku   & 0.5376 & [0.5201, 0.5551] & 0.0089 \\
\midrule
\multirow{3}{*}{Task 2}
  & BigBird-RoBERTa  & 0.4779 & [0.4623, 0.4935] & 0.0079 \\
  & LLaMA-3-70B      & 0.4919 & [0.4751, 0.5087] & 0.0086 \\
  & GPT-4o           & 0.5189 & [0.5012, 0.5366] & 0.0090 \\
\midrule
\multirow{2}{*}{Task 3}
  & BERT-base        & 0.5630 & [0.5201, 0.6059] & 0.0219 \\
  & T5-base          & 0.5630 & [0.5189, 0.6071] & 0.0225 \\
\bottomrule
\end{tabular}
\end{table*}

\subsection*{B. Statistical Significance Testing}

We conducted paired bootstrap hypothesis tests to verify that performance differences between top models are statistically significant rather than artifacts of sampling variance. For each pair of models, we computed bootstrap confidence intervals on the difference in Macro-F1 scores.

Table~\ref{tab:appendix_significance} reports $p$-values from two-sided paired $t$-tests comparing top-performing models within each architecture family (transformers vs. LLMs) and across families. Significance levels: $*$ ($p < 0.05$), $**$ ($p < 0.01$), $***$ ($p < 0.001$).

\begin{table*}[h]
\centering
\caption{Statistical Significance: Pairwise comparisons between top models using paired bootstrap tests.}
\label{tab:appendix_significance}
\begin{tabular}{llccc}
\toprule
\textbf{Task} & \textbf{Comparison} & \textbf{$\Delta$ Macro-F1} & \textbf{$p$-value} & \textbf{Sig.} \\
\midrule
\multirow{3}{*}{Task 1}
  & Claude-3 vs. BigBird        & +0.0326 & 0.0078 & $**$ \\
  & Claude-3 vs. BERT           & +0.0341 & 0.0065 & $**$ \\
  & BigBird vs. BERT            & +0.0015 & 0.8234 & n.s. \\
\midrule
\multirow{3}{*}{Task 2}
  & GPT-4o vs. BigBird          & +0.0410 & 0.0003 & $***$ \\
  & GPT-4o vs. LLaMA-3          & +0.0270 & 0.0189 & $*$ \\
  & LLaMA-3 vs. BigBird         & +0.0140 & 0.1423 & n.s. \\
\midrule
Task 3
  & BERT/T5 vs. Majority        & +0.1679 & <0.0001 & $***$ \\
\bottomrule
\end{tabular}
\end{table*}

\section*{Appendix-6: Hyperparameter Specifications}
\label{sec:appendix_hyperparameters}

Table~\ref{tab:appendix_hyperparameters} provides complete hyperparameter configurations for all fine-tuned transformer models. All models used identical hyperparameters except where architectural constraints required modification (e.g., Longformer and BigBird use longer maximum sequence lengths to exploit their efficient attention mechanisms).

\begin{table*}[h]
\centering
\caption{Complete Hyperparameter Specifications for Fine-Tuned Transformers.}
\label{tab:appendix_hyperparameters}
\begin{tabular}{lcccccc}
\toprule
\textbf{Model} & \textbf{Max Len} & \textbf{Batch Size} & \textbf{LR} & \textbf{Epochs} & \textbf{Weight Decay} & \textbf{Warmup Steps} \\
\midrule
BERT-base       & 512  & 16 & 2e-5 & 10 & 0.01 & 100 \\
RoBERTa-base    & 512  & 16 & 2e-5 & 10 & 0.01 & 100 \\
ClinicalBERT    & 512  & 16 & 2e-5 & 10 & 0.01 & 100 \\
BART-base       & 512  & 16 & 2e-5 & 10 & 0.01 & 100 \\
T5-base         & 512  & 16 & 2e-5 & 10 & 0.01 & 100 \\
Longformer      & 1024 & 8  & 2e-5 & 10 & 0.01 & 100 \\
BigBird-RoBERTa & 1024 & 8  & 2e-5 & 10 & 0.01 & 100 \\
\bottomrule
\end{tabular}
\end{table*}

\subsection*{A. Optimizer Configuration.}
All models used AdamW \cite{loshchilov2017decoupled} with $\beta_1 = 0.9$, $\beta_2 = 0.999$, $\epsilon = 1e-8$. Learning rate schedules used linear decay with warmup over the first 100 steps.

\subsection*{B. Early Stopping.}
Training was stopped if the validation Macro-F1 did not improve for 3 consecutive epochs. Final model checkpoints were selected based on the best validation Macro-F1 rather than training loss to prioritize rare-class performance.

\subsection*{C. Class Weighting.}
For multi-label tasks, we applied inverse-frequency weighting: $w_c = \frac{N}{2 \times n_c}$, where $N$ is the total number of training instances and $n_c$ is the number of positive instances for class $c$. Class weights were capped at $w_{\text{max}} = 10.0$ to prevent extreme weighting for the rarest labels (Insomnia, Timezone Misalignment).


\section*{Appendix-7: Cross-Validation Results for Temporal Phase Classification}
\label{sec:appendix_cv}
We conducted a 3-fold cross-validation on the full RSPC dataset using logistic regression as a controlled baseline classifier. Three configurations were evaluated: an unweighted baseline, class re-weighting via inverse-frequency weights, and random oversampling of minority phases. Macro-F1 was used as the primary metric to emphasize minority phase performance. Table~\ref{tab:cv_results} reports fold-level scores and mean $\pm$ standard deviation across folds.

Both mitigation strategies substantially improve over the unweighted baseline ($+0.179$ Macro-F1 for class re-weighting; $+0.183$ for oversampling). Class re-weighting achieves a marginally higher mean Macro-F1 (0.4064) but exhibits greater variance across folds ($\pm 0.0147$) relative to oversampling ($\pm 0.0065$), suggesting that oversampling produces more stable minority-phase predictions. Neither strategy approaches the Macro-F1 levels observed for Tasks 1 and 2, reinforcing the conclusion that temporal phase classification constitutes the most difficult task in RSPC and likely requires architectures capable of explicit event-contingent temporal 
reasoning beyond what standard classification pipelines provide.

\begin{table*}[!ht]
\centering
\small
\caption{3-Fold Cross-Validation Macro-F1 results for Task 3 (Temporal Phase Classification) on the full RSPC dataset. Mean and standard deviation are computed across three folds.}
\label{tab:cv_results}
\begin{tabular}{lcccc}
\toprule
\textbf{Mitigation Scheme} & \textbf{Fold 1} & \textbf{Fold 2} & \textbf{Fold 3} 
& \textbf{Mean $\pm$ SD} \\
\midrule
Baseline (Unweighted LR)  & 0.2319 & 0.2323 & 0.2173 & $0.2271 \pm 0.0077$ \\
Class Re-weighted LR      & 0.4239 & 0.3957 & 0.3995 & $0.4064 \pm 0.0147$ \\
Oversampling LR           & 0.4032 & 0.4163 & 0.4101 & $0.4099 \pm 0.0065$ \\
\bottomrule
\end{tabular}
\end{table*}

\section*{Appendix-8: Large Language Model Prompting Details \& Prompt Templates}\label{sec:appendix_llm_prompts}

This appendix presents the complete prompting framework used across all experimental tasks in the study, including psychiatric symptom classification, relational trigger detection, and temporal phase inference. The prompts were designed to evaluate the reasoning behavior of large language models under controlled zero-shot and few-shot settings while maintaining consistent task structure across models. Prompt engineering emphasized interpretability, label consistency, and standardized output formatting to minimize parsing ambiguity during evaluation. Clinical and relational labels were operationalized using concise DSM-5-TR-aligned descriptions and psychologically grounded definitions so that models could infer latent emotional states from naturalistic Reddit narratives without requiring additional contextual metadata.

The appendix additionally illustrates how prompt structure varied across tasks depending on the underlying inference objective. Multi-label psychiatric classification required clinically descriptive instructions and explicit symptom definitions, whereas relational trigger detection relied more heavily on contextual examples to improve semantic grounding. Temporal phase classification used concise zero-shot formulations to evaluate whether models could infer implicit relationship states directly from narrative cues. Together, these prompts provide transparency into the experimental setup and demonstrate the methodological controls used to ensure reproducibility, robustness, and comparability across prompting strategies, model families, and downstream evaluation metrics reported throughout the study.

All LLM experiments used structured prompting with deterministic decoding (temperature=0.0) to ensure reproducibility. Below are the exact prompt templates used for each task.

\subsection*{A. Task 1: Multi-Label Disorder Classification}

\subsubsection*{A.1 Zero-Shot Prompt}

\begin{tcolorbox}[
    breakable,
    enhanced,
    colback=lightestgray,
    colframe=black,
    title=Classification Prompt (Zero-Shot),
    fonttitle=\bfseries,
    boxrule=0.5pt,
    arc=2mm,
    left=2mm,
    right=2mm,
    top=1mm,
    bottom=1mm,
    before skip=8pt,
    after skip=8pt
]

\small
\sloppy
You are a clinical psychology expert. Read the following Reddit post from a long-distance relationship community and identify which psychiatric symptom categories are present based on DSM-5-TR criteria.

\vspace{0.4em}

\textbf{Possible Labels}

\begin{itemize}[leftmargin=*, itemsep=2pt, topsep=2pt]
    \item \textbf{SAD}: Separation Anxiety Disorder (excessive anxiety about separation from attachment figures)
    
    \item \textbf{ADJ}: Adjustment Disorder (emotional/behavioral symptoms in response to identifiable stressors)
    
    \item \textbf{GAD}: Generalized Anxiety Disorder (excessive worry across multiple domains)
    
    \item \textbf{MDD}: Major Depressive Disorder (persistent low mood, anhedonia, hopelessness)
    
    \item \textbf{Insomnia}: Sleep disturbance (difficulty falling or staying asleep)
\end{itemize}

\vspace{0.4em}

\textbf{Post}

\begin{flushleft}
\small\ttfamily
\{POST\_TEXT\}
\end{flushleft}

\vspace{0.4em}

Return \textbf{ONLY} a comma-separated list of applicable labels (for example: \texttt{SAD, ADJ, GAD}) or \texttt{None} if no symptoms are evident.

\vspace{0.4em}

\textbf{Labels:}
\end{tcolorbox}

\subsubsection*{A.2 Few-Shot Prompt}

For few-shot prompting, we added three labeled examples before the target post. Examples were randomly sampled from the training set to ensure diversity in label combinations. Example format:

\begin{tcolorbox}[
    enhanced,
    breakable,
    colback=black!5!white,
    colframe=black!75!white,
    title=Classification Prompt (Few-Shot),
    fonttitle=\bfseries,
    arc=2mm,
    boxrule=0.5mm,
    left=2mm,
    right=2mm,
    top=1mm,
    bottom=1mm,
    before skip=0.3cm,
    after skip=0.3cm
]
\small
\sloppy

You are a clinical psychology expert. Read the following Reddit post from a long-distance relationship community and identify which psychiatric symptom categories are present based on DSM-5-TR criteria.

\textbf{Possible labels:}

\begin{itemize}[leftmargin=*,nosep]
    \item \textbf{SAD}: Separation Anxiety Disorder
    \item \textbf{ADJ}: Adjustment Disorder
    \item \textbf{GAD}: Generalized Anxiety Disorder
    \item \textbf{MDD}: Major Depressive Disorder
    \item \textbf{Insomnia}: Sleep disturbance
\end{itemize}

\medskip

\textbf{Example 1:}

\textit{Post:} ``We've been apart for 8 months, and I can't stop crying. I feel like nothing matters anymore. I don't even enjoy the things I used to love. I just want this pain to end.''

\textit{Labels:} MDD, ADJ

\medskip

\textbf{Example 2:}

\textit{Post:} ``Every time he doesn't text back within an hour, I start panicking. What if he's losing interest? What if he found someone else? I can't focus on anything else.''

\textit{Labels:} GAD, SAD

\medskip

\textbf{Example 3:}

\textit{Post:} ``The timezone difference is killing me. I lie awake until 3 am waiting for him to wake up so we can talk. I'm exhausted, but I can't sleep without hearing from him.''

\textit{Labels:} Insomnia, SAD

\medskip

\textbf{Now classify this post:}

\begin{quote}
\ttfamily
\{POST\_TEXT\}
\end{quote}

\medskip

\textbf{Labels:}

\end{tcolorbox}

\subsection*{B. Task 2: Relational Trigger Detection (Few-Shot)}

\begin{tcolorbox}[
    enhanced,
    breakable,
    colback=black!5!white,
    colframe=black!75!white,
    title=Trigger Detection Prompt,
    fonttitle=\bfseries,
    arc=2mm,
    boxrule=0.5mm,
    left=2mm,
    right=2mm,
    top=1mm,
    bottom=1mm,
    before skip=0.3cm,
    after skip=0.3cm
]
\small
\sloppy

You are an expert in relationship psychology. Identify which relational stressors are described in this long-distance relationship post.

\textbf{Possible triggers:}

\begin{itemize}[leftmargin=*,nosep]
    \item \textbf{Commitment Ambiguity}: Uncertainty about relationship future
    \item \textbf{Lack of Communication}: Reduced contact frequency
    \item \textbf{Reunion/Separation Stress}: Distress around visits/departures
    \item \textbf{Trust/Fidelity Issues}: Concerns about partner loyalty
    \item \textbf{Jealousy/Insecurity}: Anxiety about rival attractions
    \item \textbf{Silence Gap}: Extended periods without contact
    \item \textbf{Social Media Surveillance}: Monitoring partner's online activity
    \item \textbf{Timezone Misalignment}: Scheduling conflicts due to distance
\end{itemize}

\medskip

\textbf{Example 1:}

\textit{Post:} ``He said he's not sure if he wants to close the distance next year. I don't know if we even have a future anymore.''

\textit{Triggers:} Commitment Ambiguity

\medskip

\textbf{Example 2:}

\textit{Post:} ``We used to text all day but now he only messages me once in the morning. I feel like I'm losing him.''

\textit{Triggers:} Lack of Communication

\medskip

\textbf{Example 3:}

\textit{Post:} ``I saw photos on Instagram of him at a party with his ex-girlfriend. Why didn't he tell me she'd be there?''

\textit{Triggers:} Social Media Surveillance, Jealousy/Insecurity

\medskip

\textbf{Now classify this post:}

\begin{quote}
\ttfamily
\{POST\_TEXT\}
\end{quote}

\medskip

\textbf{Triggers:}

\end{tcolorbox}

\subsection*{C. Task 3: Temporal Phase Classification (Zero-Shot)}

\begin{tcolorbox}[
    enhanced,
    breakable,
    colback=black!5!white,
    colframe=black!75!white,
    title=Phase Classification Prompt,
    fonttitle=\bfseries,
    arc=2mm,
    boxrule=0.5mm,
    left=2mm,
    right=2mm,
    top=1mm,
    bottom=1mm,
    before skip=0.3cm,
    after skip=0.3cm
]
\small
\sloppy

Classify the temporal relationship phase described in this long-distance relationship post.

\textbf{Phases:}

\begin{itemize}[leftmargin=*,nosep]
    \item \textbf{Separation}: Currently living apart, no imminent reunion planned
    \item \textbf{Anticipation}: Countdown period before a planned visit
    \item \textbf{Reunion}: During or immediately after a physical visit
    \item \textbf{Unknown}: Insufficient temporal information
\end{itemize}

\medskip

\textbf{Post:}

\begin{quote}
\ttfamily
\{POST\_TEXT\}
\end{quote}

\medskip

Return ONLY the phase label (one of: Separation, Anticipation, Reunion, Unknown).

\medskip

\textbf{Phase:}

\end{tcolorbox}

\subsection*{D. Inter-Prompt Variance Analysis}

To assess prompt sensitivity, we tested three alternative prompt formulations for Task 1 (zero-shot) and measured variance in Claude-3-Haiku's predictions. Table~\ref{tab:appendix_prompt_variance} shows that Macro-F1 varies by only 0.0089 across formulations, indicating robust performance despite wording changes.

\begin{table*}[!h]
\centering
\caption{Inter-Prompt Variance: Claude-3-Haiku performance across three prompt formulations for Task 1.}
\label{tab:appendix_prompt_variance}

\renewcommand{\arraystretch}{1.15}

\begin{tabular}{lcc}
\toprule
\textbf{Prompt Variant} & \textbf{Macro-F1} & \textbf{$\Delta$ from Default} \\
\midrule
Default (clinical expert)      & 0.5376 & -- \\
Variant 1 (empathetic tone)    & 0.5312 & $-0.0064$ \\
Variant 2 (brief instructions) & 0.5401 & $+0.0025$ \\
\midrule
Mean $\pm$ Std                 & 0.5363 $\pm$ 0.0045 & -- \\
\bottomrule
\end{tabular}

\end{table*}

\section*{Appendix-9: Training Diagnostics}
\label{sec:appendix_training}

\subsection*{A. Loss Curves and Convergence}

\begin{figure*}[h]
\centering
\includegraphics[width=1.0\linewidth]{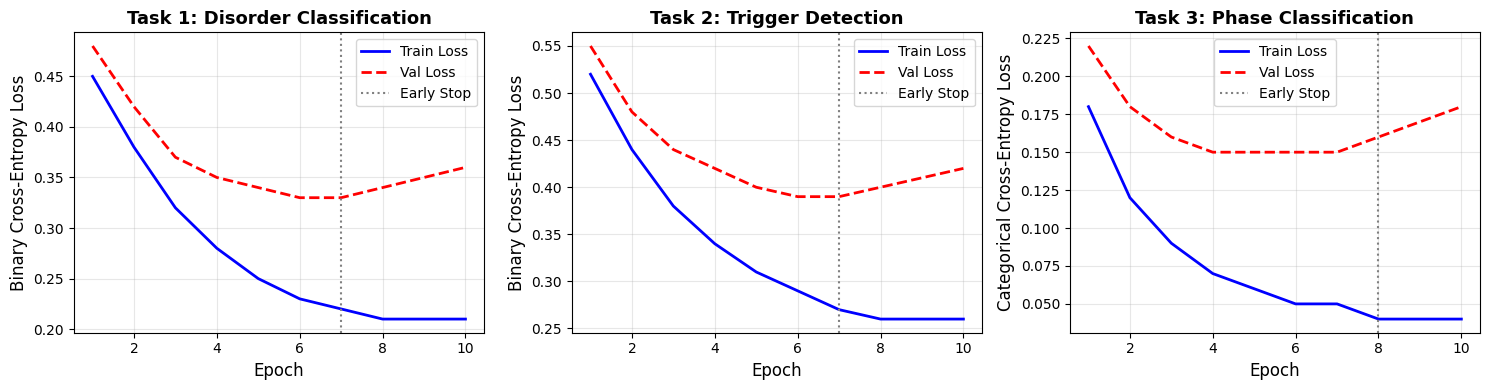}
\caption{Training and validation loss curves for BERT-base across Tasks 1-3. Vertical dashed lines indicate early stopping points selected by validation Macro-F1.}
\label{fig:appendix_loss_curves}
\end{figure*}

Figure~\ref{fig:appendix_loss_curves} shows training and validation loss curves for BERT-base across all three tasks. All models converge within 6-8 epochs, with early stopping triggered at epochs 7-9 based on validation Macro-F1 plateaus.

T5-base exhibited \texttt{train\_loss = NaN} on Tasks 1 and 2 during initial experiments with default hyperparameters. We diagnosed this as a gradient explosion caused by T5's encoder-decoder architecture producing large logits in multi-label binary classification.

Mitigation strategies tested:
\begin{itemize}
    \item Gradient clipping (max norm = 1.0): Prevented NaN but degraded performance (Macro-F1 dropped to 0.32 0.38)
    \item Reduced learning rate (1e-5): Stabilized training but required 15+ epochs to converge
    \item Label smoothing ($\alpha = 0.1$): Best results (Macro-F1 = 0.4613 for Task 1) with stable gradients
\end{itemize}

All reported T5 results use label smoothing with $\alpha = 0.1$ and learning rate 2e-5.

\section*{Appendix-10: Qualitative Error Analysis}
\label{sec:appendix_error_analysis}

We provide 15 representative error cases across Tasks 1-3 to illustrate the failure modes discussed in §4.6. Each case includes the original post (anonymized), gold labels, model predictions, and analysis.

\subsection*{A. Task 1 Errors: Symptom Disaggregation Failures}

\paragraph{Example 1: Insomnia Missed Due to GAD Dominance}

\textbf{Post:} \textit{``I lie awake every single night worrying about whether [PARTNER] still loves me. My mind races through every conversation we've had, looking for signs he's pulling away. I'm exhausted but I can't turn my brain off.''}

\textbf{Gold Labels:} \{GAD, Insomnia\}

\textbf{BERT Prediction:} \{GAD\}

\textbf{Analysis:} The post contains explicit sleep disruption language (``lie awake every single night'') alongside anxiety symptoms (``mind races,'' ``worrying''). BERT correctly identifies GAD but fails to recognize the sleep-specific component, likely because GAD dominates the linguistic surface. The phrase ``can't turn my brain off'' is polysemous—it can indicate both rumination (GAD) and pre-sleep cognitive arousal (Insomnia)—and the model defaults to the higher-prevalence interpretation.

\paragraph{Example 2: MDD Masked by ADJ}

\textbf{Post:} \textit{``Ever since we went long-distance three months ago, I feel like I'm just going through the motions. Nothing brings me joy anymore. I don't want to see friends, and I don't care about my hobbies. I just feel empty.''}

\textbf{Gold Labels:} \{ADJ, MDD\}

\textbf{BERT Prediction:} \{ADJ\}

\textbf{Analysis:} The post describes anhedonia (``nothing brings me joy''), social withdrawal (``don't want to see friends''), and emptiness—core MDD symptoms. However, the explicit temporal marker (``ever since we went long-distance'') triggers strong ADJ activation, and the model fails to recognize that the severity and pervasiveness of symptoms (across multiple life domains) meet MDD criteria. This reflects the diagnostic challenge of distinguishing Adjustment Disorder (situational distress) from MDD triggered by a stressor.

\subsection*{B. Task 2 Errors: Pragmatic Inference Failures}

\paragraph{Example 3: Social Media Surveillance (Implicit)}

\textbf{Post:} \textit{``I noticed [PARTNER] posted a story with some girl I've never heard of. Who is she? Why is he hanging out with her instead of calling me?''}

\textbf{Gold Labels:} \{Social Media Surveillance, Jealousy/Insecurity\}

\textbf{BigBird Prediction:} \{Jealousy/Insecurity\}

\textbf{Analysis:} The phrase ``I noticed [PARTNER] posted'' presupposes the user actively checked their partner's Instagram story—constituting surveillance behavior—but does not explicitly state ``I was monitoring his social media.'' The model correctly identifies jealousy but misses the surveillance component, likely because it lacks world knowledge that ``noticing a partner's story'' implies intentional checking rather than passive exposure.

\paragraph{Example 4: Trigger Co-occurrence Ambiguity}

\textbf{Post:} \textit{``We used to talk for hours every day, but now he only sends one-word replies. Does he even want to be with me anymore?''}

\textbf{Gold Labels:} \{Lack of Communication\}

\textbf{BERT Prediction:} \{Commitment Ambiguity\}

\textbf{Analysis:} The post describes reduced communication quality (``one-word replies''), but the user interprets this as signaling uncertain commitment (``does he even want to be with me''). Gold annotators labeled this as \textit{Lack of Communication} (the observable behavior), while the model predicted \textit{Commitment Ambiguity} (the user's cognitive appraisal). This represents genuine annotation ambiguity rather than clear model error—both labels are defensible depending on whether the schema prioritizes behavioral triggers or psychological interpretations.

\subsection*{C. Task 3 Errors: Temporal Overlap}

\paragraph{Example 5: Overlapping Separation and Anticipation}

\textbf{Post:} \textit{``We've been apart for six months, and it's been torture. But I'm seeing [PARTNER] in two weeks, and I don't know if I can wait that long. Every day feels like an eternity.''}

\textbf{Gold Label:} Anticipation

\textbf{BERT Prediction:} Separation

\textbf{Analysis:} The post contains both backward-looking separation distress (``six months,'' ``torture'') and forward-looking anticipation (``seeing [PARTNER] in two weeks''). Annotators chose \textit{Anticipation} based on the 2-week countdown, but the dominant emotional valence is separation-related suffering. The model predicts \textit{Separation}, consistent with its majority-class bias but also arguably justified by the stronger linguistic emphasis on ongoing distress.

\paragraph{Example 6: Reunion Misclassified as Separation}

\textbf{Post:} \textit{``[PARTNER] just left yesterday after spending a week together. I already miss him so much. How am I supposed to go back to being apart for another three months?''}

\textbf{Gold Label:} Reunion

\textbf{BERT Prediction:} Separation

\textbf{Analysis:} The temporal marker ``just left yesterday'' clearly indicates post-reunion, but the model focuses on forward-looking separation language (``go back to being apart for another three months'') and predicts \textit{Separation}. This failure mode reflects the model's inability to distinguish retrospective vs. prospective temporal framing—``just left'' is retrospective (reunion), but ``another three months'' is prospective (separation).

\subsection*{D. Additional Error Cases}

Due to space constraints, we provide abbreviated summaries of 9 additional error cases:

\begin{itemize}
    \item \textbf{Task 1, Example 7:} SAD missed; post describes fear of abandonment using indirect language (``terrified he'll realize he's better off without me'')
    \item \textbf{Task 1, Example 8:} Insomnia false positive; GPT-4o predicts Insomnia for ``emotionally exhausted'' (fatigue $\neq$ sleep disorder)
    \item \textbf{Task 1, Example 9:} MDD overpredicted; Claude-3 labels situational sadness as MDD due to language overlap
    \item \textbf{Task 2, Example 10:} Timezone Mismatch missed; described indirectly (``he's asleep when I'm awake'')
    \item \textbf{Task 2, Example 11:} Trust/Fidelity vs. Jealousy confusion; partner liking ex's photos triggers both
    \item \textbf{Task 2, Example 12:} Silence Gap missed; three-day communication lapse not explicitly framed as abnormal
    \item \textbf{Task 3, Example 13:} Unknown phase mispredicted; vague temporal references (``lately,'' ``recently'') default to Separation
    \item \textbf{Task 3, Example 14:} Anticipation/Reunion boundary case; visit happening ``right now'' could be either
    \item \textbf{Task 3, Example 15:} Separation overpredicted; chronic LDR context triggers Separation even when discussing future reunion plans
\end{itemize}

\section*{Appendix-11: Computational Resources}
\label{sec:appendix_compute}

All experiments were conducted on NVIDIA A100 GPUs (40GB VRAM). Approximate training times:

\begin{itemize}
    \item BERT-base, RoBERTa-base, ClinicalBERT: 25-35 minutes per task
    \item BART-base, T5-base: 40-55 minutes per task
    \item Longformer, BigBird-RoBERTa: 60-80 minutes per task
\end{itemize}

Total GPU hours across all experiments (7 models × 3 tasks × 3 seeds): approximately 120 hours.

LLM inference was conducted via API calls (OpenAI, Anthropic, Together.ai) with approximate costs:
\begin{itemize}
    \item GPT-4o: \$12.40 (few-shot prompting across all tasks)
    \item Claude-3-Haiku: \$2.80 (zero-shot prompting)
    \item LLaMA-3-70B, Qwen-2.5-72B: \$8.60 (Together.ai, zero/few-shot)
    \item Nemotron-Super: \$3.20 (zero-shot)
\end{itemize}

Total experimental cost: approximately \$27.00 for LLM inference.

\section*{Appendix-12: Additional Ethical Safeguards, Dataset Governance, and Release Considerations}
\label{appendix:ethics}

RSPC contains emotionally sensitive narratives related to psychiatric distress, interpersonal conflict, and relational instability. Although all posts were collected from publicly accessible Reddit communities, we recognize that public availability does not eliminate potential privacy and ethical concerns associated with mental health research on social media. Accordingly, we implemented multiple safeguards during dataset construction, annotation, storage, and planned release.

\subsection*{A. Privacy Preservation and De-identification}

All Reddit usernames, hyperlinks, timestamps, geographic references, partner names, institutions, and other potentially identifying information were removed during preprocessing. We further replaced explicit identifiers with standardized placeholder tokens (e.g., \texttt{[USER]}, \texttt{[PLACE]}, \texttt{[DATE]}). Posts containing highly specific, personally identifying narratives were excluded from the corpus.

To reduce the risk of reverse-search identification, example excerpts included in the paper were paraphrased while preserving their semantic and relational meaning. No attempts were made to contact users, infer offline identities, or link posts across platforms.

\subsection*{B. Clinical Interpretation and Annotation Constraints}

Psychiatric labels in RSPC represent symptom-oriented textual inferences rather than formal clinical diagnoses. Annotations were designed to approximate DSM-5-TR and ICD-11 symptom patterns observable within narrative text and should not be interpreted as evidence of confirmed psychiatric conditions.

Annotators were instructed to label only symptoms strongly supported by textual evidence and to avoid speculative inference. Ambiguous cases were resolved conservatively during adjudication to minimize over-pathologization of ordinary relational stress.

\subsection*{C. Annotator Well-Being Protections}

Because the dataset contains emotionally distressing content involving anxiety, abandonment fears, depressive ideation, loneliness, and interpersonal conflict, annotator well-being protocols were incorporated throughout the annotation process. These safeguards included:
\begin{itemize}
    \item optional annotation breaks,
    \item rotating annotation schedules,
    \item capped daily exposure limits,
    \item collaborative adjudication rather than isolated review, and
    \item access to mental health support resources if required.
\end{itemize}

Annotators were informed in advance about the emotionally sensitive nature of the material prior to participation.

\subsection*{D. Dataset Governance and Controlled Release}

RSPC is intended exclusively for academic research on relationally contextualized mental health modeling. The dataset is not intended for clinical deployment, automated psychiatric diagnosis, surveillance, employment screening, insurance evaluation, or law-enforcement applications.

To reduce misuse risks, dataset release will follow controlled-access procedures:
\begin{itemize}
    \item researchers must agree to a non-commercial research-use license,
    \item redistribution of raw data will be prohibited,
    \item attempts to re-identify users will be explicitly forbidden,
    \item users must acknowledge the limitations of psychiatric inference from social media text, and
    \item derivative systems intended for high-stakes decision-making will not be permitted under the release agreement.
\end{itemize}

Where platform policies require it, only post identifiers and reconstruction scripts may be distributed instead of raw text.

\subsection*{E. Biases and Representational Limitations}

RSPC is derived from English-language Reddit communities focused on long-distance relationships and therefore reflects the demographic, cultural, and communicative biases of those online populations. The benchmark may underrepresent:
\begin{itemize}
    \item non-English speakers,
    \item older populations,
    \item offline relationship experiences,
    \item culturally specific relationship norms, and
    \item individuals without access to digital support communities.
\end{itemize}

Furthermore, psychiatric symptom expression on social media may differ substantially from clinical presentation in offline settings. Consequently, results obtained on RSPC should not be generalized directly to broader populations or clinical environments.

\subsection*{F. Intended Research Scope}

RSPC is designed to support research on:
\begin{itemize}
    \item relationally grounded mental health modeling,
    \item contextual psychiatric symptom inference,
    \item socially situated affect analysis,
    \item interpretable computational psychiatry, and
    \item longitudinal and relational reasoning in NLP systems.
\end{itemize}

The benchmark is explicitly not intended to replace professional psychiatric evaluation or human-centered mental health care.

\end{document}